\documentclass[twocolumn,preprintnumbers,amsmath,amssymb,showkeys]{revtex4}
\usepackage{hyperref}
\usepackage{graphicx}
\numberwithin{equation}{section}
\pdfoutput=1

\newcommand{\beq}{\begin{eqnarray}}
\newcommand{\eeq}{\end{eqnarray}}
\newcommand{\bfig}{\begin{figure}}
\newcommand{\efig}{\end{figure}}
\newcommand{\cen}{\centering}
\newcommand{\incgraph}[1]{\includegraphics[width=.5\textwidth]{{#1}}}
\newcommand{\fref}[1]{Fig.~\ref{#1}}

\newcommand{\A}{\mathcal{A}}
\newcommand{\C}{\mathcal{C}}
\newcommand{\G}{\mathcal{G}}
\newcommand{\N}{\mathcal{N}}
\newcommand{\E}{\mathcal{E}}
\newcommand{\El}{\mathcal{E}_{1}}
\newcommand{\Es}{\mathcal{E}_{0}}
\newcommand{\Eml}{\mathcal{E}_{\infty}}
\newcommand{\lap}{\mathbf{\Delta}}
\newcommand{\bh}{\textbf{h}}
\newcommand{\bbf}{\textbf{f}}
\newcommand{\bA}{\mathbf{A}}
\newcommand{\bD}{\mathbf{D}}
\newcommand{\bE}{\mathbf{E}}
\newcommand{\bG}{\mathbf{G}}
\newcommand{\bP}{\mathbf{P}}
\newcommand{\bQ}{\mathbf{Q}}
\newcommand{\bV}{\mathbf{V}}
\newcommand{\bS}{\mathbf{V}_{+}}
\newcommand{\bT}{\mathbf{V}_{-}}
\newcommand{\bbT}{\mathbf{T}}
\newcommand{\quart}{\frac{1}{4}}
\newcommand{\half}{\frac{1}{2}}
\newcommand{\Iyz}{I\left [ Y;Z \right ] }
\newcommand{\Ixz}{I\left [ X;Z \right ] }
\newcommand{\ft}{\tilde{t}}
\newcommand{\lhr}{\langle \bh \rangle}
\newcommand{\lhyr}{\langle \bh | y \rangle}
\newcommand{\lhhr}{\langle h_x h_{x'} \rangle_{2t}}
\newcommand{\sxy}{\sum_{x,y=1}^n}

\newcommand{\shxym}{\sum_{h_x \times h_y=-1}}
\newcommand{\shxp}{\displaystyle\sum_{h_x=+1}}
\newcommand{\shxm}{\displaystyle\sum_{h_x=-1}}
\newcommand{\sx}{\sum_{x}}
\newcommand{\dprime}{{\prime\prime}}
\newcommand{\sxd}{\sum_{x^{\dprime}}}
\newcommand{\sy}{\sum_{y=1}^n}
\newcommand{\syz}{\sum_{y,z}}
\newcommand{\sz}{\sum_{z=1}^K}
\newcommand{\sxx}{\sum_{x,x'=1}^n}
\newcommand{\expy}[1]{\left\langle #1 \right\rangle_y}
\newcommand{\expz}[1]{\left\langle #1 \right\rangle_z}
\renewcommand{\exp}[1]{\left\langle #1 \right\rangle}

\newcommand{\spincgraph}[1]{\includegraphics[width=\textwidth]{{#1}}}

\begin{document}

\title{An information-theoretic derivation of min-cut based clustering}

\author{Anil Raj}
 \affiliation{Department of Applied Physics and Applied Mathematics\\ Columbia University, New York}
 \email{ar2384@columbia.edu}
\author{Chris H. Wiggins}
 \affiliation{Department of Applied Physics and Applied Mathematics\\
                Center for Computational Biology and Bioinformatics\\ Columbia University, New York}
 \email{chris.wiggins@columbia.edu}

\date{\today}

\begin{abstract}
Min-cut clustering, based on minimizing one of two heuristic cost-functions proposed by Shi and Malik, 
has spawned tremendous research, both analytic and algorithmic, in the graph partitioning and 
image segmentation communities over the last decade. It is however unclear if these heuristics 
can be derived from a more general principle facilitating generalization
to new problem settings. Motivated by an existing graph partitioning framework,
we derive relationships between optimizing relevance information,
as defined in the Information Bottleneck method, and the regularized 
cut in a K-partitioned graph. For fast mixing graphs, we show that the cost functions 
introduced by Shi and Malik can be well approximated as the rate of loss of predictive information about the location of 
random walkers on the graph. For graphs generated from a stochastic algorithm designed 
to model community structure, the optimal information theoretic partition 
and the optimal min-cut partition are shown to be the same with high probability.

\end{abstract}


\maketitle

\renewcommand{\thesection}{\arabic{section}}
\section{Introduction}

Min-cut 
based graph partitioning has been used successfully to
find clusters in networks, with applications in image
segmentation as well as clustering biological and
sociological networks. The central idea is to develop fast and 
efficient algorithms that optimally cut the edges between graph nodes, 
resulting in a separation of graph nodes into clusters.
Particularly, since Shi and Malik successfully
showed \cite{malik} that the \emph{average} cut and the \emph{normalized} cut 
(defined below) were useful heuristics to be optimized, there has been tremendous 
research in constructing the best normalized-cut-based cost function
in the image segmentation community. 

The Information Bottleneck (IB) method \cite{bialek,bottleneck} is 
a clustering technique, 
based on rate-distortion theory \cite{shannon}, that has been 
successfully applied in a wide variety of contexts including 
clustering word documents and gene-expression profiles 
\cite{slonimthesis}. The IB method is also capable of learning
clusters in graphs and has been used successfully for 
synthetic and actual networks \cite{infomod}. In the hard clustering case, 
given the diffusive probability distribution over a graph,
IB optimally assigns probability distributions,
associated with nodes, into distinct groups. These assignment rules 
define a separation of the graph nodes into clusters.

We here illustrate how minimizing the 
two cut-based heuristics introduced by Shi and Malik can be well-approximated 
by the rate of loss of \emph{relevance information}, defined in
the IB method applied to clustering
graphs. To establish these relations, we must first define the 
graphs to be partitioned; we assume hard-clustering and the cluster 
cardinality to be $K$. We show, numerically, that maximizing mutual 
information and minimizing \emph{regularized} cut amount to the same 
partition with high probability, for more modular 32-node graphs,
where \emph{modularity} is defined by the probability of inter-cluster edge 
connections in the Stochastic Block Model for graphs 
(See \hyperref[results]{\sc{Numerical Experiments}}). 
We also show that the optimization goal of maximizing relevance information is 
equivalent to minimizing the regularized cut for 16-node graphs.\footnote{We 
chose 16-node graphs so the network and its partitions
could be parsed visually with ease.}

\section{The Min-Cut Problem}
Following \cite{uli}, for an undirected, unweighted graph $\G=(\bV,\bE)$ with 
$n$ nodes and $m$ edges, represented\footnote{We use the shorthand $x \sim y$ 
to mean $x$ is adjacent to $y$.} by its adjacency matrix 
$\bA := \{ A_{xy} = 1 \iff x \sim y \}$, 
we define for two not necessarily disjoint sets of nodes $\bS,\bT\subseteq \bV$,
the association
\beq
W(\bS,\bT) = \sum_{x\in \bS, y\in \bT} A_{xy}.
\eeq

We define a bisection of $\bV$ into $\bV_{\pm}$ if 
$\bS\cup \bT = \bV$ and $\bS\cap \bT = \emptyset$. 
For a bisection of $\bV$ into $\bS$ and $\bT$, the `cut' is defined as 
$c(\bS,\bT) = W(\bS,\bT)$. 
We also quantify the size of a set $\bS \subseteq \bV$ in terms of the number of nodes
in the set $\bS$ or the number of edges with at least one node in the set $\bS$:
\beq
\omega(\bS) & = & \sum_{x\in \bS} 1\notag \\
\Omega(\bS) & = & \sum_{x\in \bS} d_x,
\eeq
where $d_x$ is the degree of node $x$.

Shi and Malik \cite{malik} defined a pair of regularized cuts, 
for a bisection of $\bV$ into $\bS$ and $\bT$; 
the \emph{average cut} was defined as
\beq
\A =
\frac{W(\bS,\bT)}{\omega(\bS)}
+
\frac{W(\bS,\bT)}{\omega(\bT)}
\eeq
and the \emph{normalized cut} was defined as
\beq
\N =
\frac{W(\bS,\bT)}{\Omega(\bS)}
+
\frac{W(\bS,\bT)}{\Omega(\bT)}.
\eeq

This definition can be generalized, for a $K$-partition of $\bV$ into 
$\bV_1,\bV_2,...,\bV_K$ \cite{uli}, to
\beq
\A & = & \sum_j \frac{W(\bV_j,\bar{\bV}_j)}{\omega(\bV_j)} \\
\N & = & \sum_j \frac{W(\bV_j,\bar{\bV}_j)}{\Omega(\bV_j)}
\eeq
where $\bar{\bV}_j = \bV \setminus \bV_j$.

For the graph $\G$, we can define the graph Laplacian $\lap = \bD - \bA$ 
where $\bD$ is a diagonal matrix of vertex degrees.
For a bisection of $\bV$, we also define the partition indicator vector 
$\bh$ \cite{fiedler}
\beq
h_x = \left \{ \begin{array}{ll}
+1 & \mbox{$\forall x \in \bS$} \\
-1 & \mbox{$\forall x \in \bT$}. \end{array} \right.
\eeq
Specifying two `prior' probability distributions over the set of nodes $\bV$ : 
(i) $p(x) \propto 1$ and (ii) $p(x) \propto d_x$, 
we then define the \textit{average} of $\bh$ to be
\beq
\bar{\bh} & = & \frac{\sum_{x \in \bV} h_x}{n}\notag \\
\lhr & = & \frac{\sum_{x \in \bV} d_x h_x}{2m}.
\eeq

The cut, as defined by Fiedler \cite{fiedler}, and the regularized cuts,
as defined by Shi and Malik \cite{malik}, can then by written in terms of 
$\bh$ as (See \hyperref[appendix]{\sc{Appendix}})
\beq
c & = & \quart \bh^\bbT \lap \bh\notag \\
\A & = & \frac{1}{n} \frac{\bh^\bbT \lap \bh}{1-{\bar{\bh}}^2} \\
\N & = & \frac{1}{2m} \frac{\bh^\bbT \lap \bh}{1-\lhr^2}.\notag
\eeq

More generally, for a $K$-partition, we define the partition indicator matrix $\bQ$ as
\beq \label{eq:defQ}
Q_{zx} \equiv  p(z|x) = \begin{array}{ll}
1 & \mbox{$\forall x \in z$} \end{array}
\eeq
where $z \in \{ \bV_1,\bV_2,...,\bV_K \}$
and define $\bP$ as a diagonal matrix of the `prior' probability distribution over the nodes. 
The regularized cut can then be generalized as
\beq
\C = \sum_j \frac{[\bQ^\bbT \lap \bQ]_{jj}}{[\bQ^\bbT \bP \bQ]_{jj}}
\eeq
where for $p(x) \propto 1$, $\C = \A$; and for $p(x) \propto d_x$, $\C = \N$.

Inferring the optimal $\bh$ (or $\bQ$), however, has been shown to be an NP-hard combinatorial 
optimization problem \cite{wagner}. 

\section{Information Bottleneck}
Rate-distortion theory, which provides the foundations for lossy data compression, formulates 
clustering in terms of a compression problem; it determines the code with minimum average length
such that information can be transmitted without exceeding some
specified distortion. Here, the model-complexity, or \emph{rate}, is measured by the mutual information 
between the data and their representative codewords (average number of bits used to store a data 
point). Simpler models correspond to smaller rates but they typically suffer from relatively 
high \emph{distortion}. The distortion measure, which can be identified
with loss functions, usually depends on the problem; in the simplest of cases, it is
the variance of the difference between data and their representatives. 

The Information Bottleneck (IB) method \cite{bottleneck} proposes the use of mutual information 
as a natural distortion measure. In this method, the data are 
compressed into clusters while maximizing the amount of information
that the `cluster representation' preserves about some specified \emph{relevance} variable. For example, 
in clustering word documents, one could use the `topic' of a document as the relevance variable. 

For a graph $\G$, let $X$ be a random variable over graph nodes, $Y$ be the relevance variable 
and $Z$ be the random variable over clusters. 
Graph partitioning using the IB method \cite{infomod} 
learns a probabilistic cluster assignment function $p(z|x)$ which gives the probability that a given
node $x$ belongs to cluster $z$. The optimal $p(z|x)$ minimizes the mutual information between
$X$ and $Z$, while minimizing the loss of predictive information between $Z$ and $Y$. This 
complexity--fidelity trade-off can be expressed in terms of a functional to be minimized
\beq
\mathcal{F}[p(z|x)] = -\Iyz + T \Ixz
\eeq
where the temperature $T$ parameterizes the relative importance of precision over complexity. 
As $T \rightarrow 0$, 
we reach the `hard clustering' limit where each node is assigned with unit probability
to one cluster (i.e $p(z|x) \in \{0,1\}$).

Graph clustering, as formulated in terms of the IB method, requires a joint distribution 
$p(y,x)$ to be defined on the graph; we use the distribution given by continuous 
graph diffusion as it naturally captures topological information about the 
network \cite{infomod}. The relevance variable $Y$ then ranges over the nodes 
of the graph and is defined as the node at which a random walker ends at time $t$ if 
the random walker starts at node $x$ at time $0$. For continuous time diffusion, 
the conditional distribution $p^t(y|x)$ is given as
\beq
\bG^t = p^t(y|x) = e^{- t\lap\bP^{-1}} 
\eeq
where $\lap$ is the graph Laplacian and $\bP$ a diagonal matrix of the prior 
distribution over the graph nodes, as described earlier. The 
characteristic diffusion time scale $\tau$ of the system is given by the inverse of the smallest
non-zero eigenvalue of the diffusion operator exponent $\lap \bP^{-1}$ and characterizes the slowest 
decaying mode in the system. To calculate the joint distribution $p(y,x)$ from the conditional $\bG^t$, 
we must specify an initial or prior distribution\footnote{Strictly speaking, any diagonal matrix $\bP$ that we specify
determines the steady-state distribution. Since we are modeling the distribution of random walkers at statistical
equilibrium, we always use this distribution as our initial or prior distribution.}; 
we use the two different priors $p(x)$, used earlier to calculate the expected value of $\bh$ : 
(i) $p(x) \propto 1$ and (ii) $p(x) \propto d_x$.

\section{Rate of Information Loss in Graph Diffusion}
We analyze here the rate of loss of predictive information 
between the relevance variable $Y$ and the cluster
variable $Z$, during diffusion on a graph $\G$, 
after the graph nodes have been hard-partitioned into K clusters.
\subsection{Well-mixed limit of graph diffusion}
For a given partition $\bQ$ of the graph, defined in Eqn. \eqref{eq:defQ}, 
we approximate the mutual information $\Iyz$ when diffusion on the graph reaches its well-mixed limit. 
We introduce the \textit{dependence} $\eta(y,z)$ such that
\beq
p(y,z) = p(y)p(z)(1+\eta).
\eeq
This implies $\expy{\eta} = \expz{\eta} = 0$ and
$\expy{\expz{\eta^2}} = \exp{\eta}$ where $\exp{}$ denotes expectation over the joint distribution
and $\expy{}$ and $\expz{}$ denote expectation over the corresponding marginals.

In the well-mixed limit, we have $\eta \ll 1$.
The predictive information (expressed in nats) can then be approximated as:
\beq
\Iyz & = & \exp{\ln{\frac{p(z,y)}{p(z)p(y)}}}\notag \\ 
& = & \expz{\expy{(1 + \eta) \ln{(1+\eta)}}}\notag \\
& \approx & \expz{\expy{(1 + \eta) (\eta - \half \eta^2)}}\notag \\
& \approx & \expz{\expy{\eta + \half \eta^2}}\notag \\ 
& = & \half \expz{\expy{\eta^2}} \\
& = & \half \syz p(y) p(z) \left( \frac{p(z,y)}{p(z)p(y)} -1 \right)^2\notag \\
& = & \half \left( \syz \frac{p(y,z)^2}{p(y)p(z)} -1 \right) \equiv \iota.
\eeq
Here, we define $\iota$ as a first-order approximation to $\Iyz$ in the well-mixed limit of graph diffusion.
\subsubsection{Well-mixed $K$-partitioned graph}
As in the IB method, the Markov condition $Z-X-Y$ allows us to make several simplifications
for the conditional distributions and associated information theoretic measures.
For a $K$-partition $\bQ$ of the graph, we have
\beq
p(y,z) & = & \sx p(x,y,z)\notag \\
& = & \sx p(z|y,x) p(y|x) p(x)\notag \\
& = & \sx p(z|x) p(y|x) p(x) \equiv \bQ\bP{\bG^t}^\bbT. \\
p(y,z)^2 & = & \left( \sx p(z|x) p(y|x) p(x) \right)^2\notag \\
& = & \sxx p(z|x) p(y|x) p(x) p(z|x') p(y|x') p(x')\notag \\
& = & \sxx Q_{zx} G^t_{yx} P_x Q_{zx'} G^t_{yx'} P_{x'}. \\
p(z) & = & \sx p(z|x) p(x)\notag \\
& = & \sx Q_{zx} P_{x}.
\eeq
Graph diffusion being a Markov process, we have 
$\sy G^t_{x'y} G^t_{yx} = G^{2t}_{x'x}$. Using this and Bayes rule 
$G^t_{yx} P_x = G^t_{xy} P_y$, we have
\beq
\iota & = & \half \left( \syz \frac{\sxx Q_{zx} G^t_{yx} P_x Q_{zx'} G^t_{yx'} P_{x'}}
{(\sxd Q_{zx^\dprime} P_{x^\dprime}) P_y} - 1 \right)\notag \\
& = & \half \left( \syz \frac{\sxx Q_{zx} Q_{zx'} P_y G^t_{x'y} G^t_{yx} P_{x}}
{(\sxd Q_{zx^\dprime} P_{x^\dprime}) P_y} - 1 \right)\notag \\
& = & \half \left( \sz \frac{\sxx Q_{zx} Q_{zx'} (\sy G^t_{x'y} G^t_{yx}) P_{x}}
{(\sxd Q_{zx^\dprime} P_{x^\dprime})} - 1 \right)\notag \\
& = & \half \left( \sz \frac{\sxx Q_{zx} Q_{zx'} G^{2t}_{x'x} P_{x}}
{(\sxd Q_{zx^\dprime} P_{x^\dprime})} - 1 \right).
\eeq
In the hard clustering case,  
$\sx Q_{zx} P_{x}=p(z)=[\bQ\bP\bQ^\bbT]_{zz}$ and we have
\beq
\iota & = & \half \left( \sz \frac{[\bQ(\bG^{2t}\bP)\bQ^\bbT]_{zz}}{[\bQ\bP\bQ^\bbT]_{zz}} - 1 \right).
\eeq
\subsubsection{Well-mixed 2-partitioned graph}
We can re-write $\iota$ as
\beq
\iota & = & \half \expz{\expy{\eta^2}}\notag \\
& = & \half \expy{\expz{\frac{(p(z|y)-p(z))^2}{p(z)^2}}}.
\eeq
For a bisection $\bh$ of the graph, $z \in \{ +1,-1 \}$ and we have
\beq
p(z|x) & = & \half (1 \pm h_x) \equiv \half (1+zh_x). \\
p(z|y) & = & \frac{1}{p(y)} \sx p(z,y,x)\notag \\
& = & \frac{1}{p(y)} \sx p(z|x) p(y|x) p(x)\notag \\
& = & \half \sx (1+zh_x) p(x|y)\notag \\
& = & \half (1+z\lhyr). \\
p(z) & = & \sx p(z,x) = \sx p(z|x) p(x)\notag \\
& = & \half \sx (1+zh_x) p(x)\notag \\
& = & \half (1+z\lhr). \\
p(z|y)-p(z) & = & \half (1+z\lhyr) - \half (1+z\lhr)\notag \\
& = & \half z(\lhyr-\lhr).
\eeq
We then have
\beq
\expz{\frac{(p(z|y)-p(z))^2}{p(z)^2}} & = & \sz \frac{\quart (\lhyr-\lhr)^2}{\half (1+z\lhr)}\notag \\
& = & \frac{(\lhyr-\lhr)^2}{2} \sz \frac{1}{1+z\lhr}\notag \\
& = & \frac{(\lhyr-\lhr)^2}{1-\lhr^2}.
\eeq
The mutual information $\Iyz$ can then be approximated as
\beq
\iota & = & \half \frac{\expy{(\lhyr-\lhr)^2}}{1-\lhr^2}\notag \\
& = & \half \frac{\sigma^2_y\left( \lhyr \right)}{1 - \lhr^2}.
\eeq
Using Bayes rule $p^t(x|y) p(y) = p^t(y|x) p(x)$, we have
\beq
\lhyr & = & \sx h_x p^t(x|y) = \sx \frac{h_x p^t(y|x) p(x)}{p(y)}.\\
\expy{\lhyr^2} & = & \sy p(y) \sxx h_x h_{x'} \frac{p^t(y|x) p(x) p^t(x'|y)}{p(y)}\notag \\
& = & \sy \sxx h_x h_{x'} p^t(y|x) p^t(x'|y) p(x).
\eeq
Again, graph diffusion being a Markov process,
\beq
\expy{\lhyr^2} & = & \sxx h_x h_{x'} p^{2t}(x'|x) p(x)\notag \\
& = & \lhhr. \\
\sigma^2\left( \lhyr \right) & = & \expy{\lhyr^2}-\lhr^2\notag \\
& = & \lhhr - \lhr^2. \\
\iota & = & \half \frac{\lhhr - \lhr^2}{1 - \lhr^2}.
\eeq

\subsection{Fast-mixing graphs}
When diffusion on a graph reaches its well-mixed limit in short times,
we have $\bG^{2t} \approx \textbf{I} - 2t\lap\bP^{-1}$. Thus, for a $K$-partition of a graph
\beq
\bQ(\bG^{2t}\bP)\bQ^\bbT & \approx & \bQ(\bP-2t\lap)\bQ^\bbT\notag \\
& = & \bQ\bP\bQ^\bbT - 2t \bQ\lap\bQ^\bbT.
\eeq
For bisections, the short-time approximation of $\lhhr$ can be written as
\beq
\lhhr & = & \sxx h_{x'} p^{2t}(x',x) h_x\notag \\
& = & \bh^\bbT \bG^{2t} \bP \bh\notag \\
& \approx & \bh^\bbT (I - 2t\lap\bP^{-1}) \bP \bh\notag \\
& = & \bh^\bbT \bP \bh - 2t \bh^\bbT \lap \bh\notag \\
& = & 1 - 2t \bh^\bbT \lap \bh.
\eeq

For fast-mixing graphs, the long-time and short-time approximations 
for $\Iyz$ and $\lhhr$, respectively, hold simultaneously.
\beq
\begin{array}{rcl}
\Iyz \approx &\iota& \approx \left( \half - t \frac{\bh^T \lap \bh}{1 - \lhr^2} \right) \\
\Rightarrow \frac{d\Iyz}{dt} \approx &\frac{d\iota}{dt}& \propto \left\{ \begin{array}{lll}
\A & ; & \mbox{$p(x) \propto 1$} \\
\N & ; & \mbox{$p(x) \propto d_x$}. \end{array} \right.
\end{array}
\eeq

We have shown analytically that, for fast mixing graphs, the heuristics introduced by Shi and Malik are 
proportional to the rate of loss of relevance information. The error incurred in the approximations 
$\Iyz \approx \iota$ and $\lhhr \approx 1-2t \bh^\bbT \lap \bh$ can be defined as
\beq
\Es(t) & = & \left| \frac{\lhhr - (1-2t \bh^\bbT \lap \bh)}{\lhhr} \right| \\
\El(t) & = & \left| \frac{\Iyz(t) - \iota(t)}{\Iyz(t)} \right|.
\eeq

\section{Numerical Experiments}
\label{results}
The validity of the two approximations can be seen in a typical plot 
of $\El(t)$ and $\Es(t)$ as a function of normalized diffusion time $\ft=t/\tau$,
for the two different choices of prior distributions over the nodes. 
$\El$, as seen in \fref{plotone}, 
is often found to be non-monotonic and sometimes exhibits oscillations.
This suggests defining $\Eml$, a modified monotonic `$\El$':
\beq
\Eml(t) \equiv \max_{t' \geq t} \El(t').
\eeq
We don't need to define a monotonic form for $\Es$ since this error is always found to be
monotonically increasing in time.

\bfig
\cen
\incgraph{plotone}
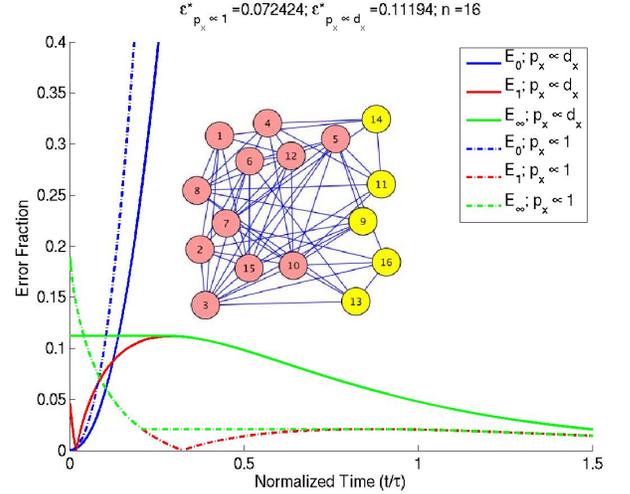
\caption{$\El$ and $\Es$ vs normalized diffusion time for two choices of priors over
the graph nodes. $\El$ (red) typically tends to have a non-monotonic behavior which
motivates defining a monotonic $\Eml$ (green). \label{plotone}}
\efig

By fast-mixing graphs, we mean graphs which become well-mixed in short times, 
i.e. graphs for which both the long-time and short-time approximations 
hold simultaneously within a certain range of time 
$\ft^*_- \leq \ft \leq \ft^*_+$, as illustrated in \fref{plotone}, 
where we define
\beq
\E(t) & = & \max(\Eml(t),\Es(t)) \\
\E^* & = & \min_{t} \E(t) \\
\ft^*_- & = & \min(\arg\min_{\ft} \E(\ft) ) \\
\ft^*_+ & = & \max(\arg\min_{\ft} \E(\ft) ).
\eeq
Note that the use of $\Eml$ instead of $\El$ over-estimates the value of $\E^*$;
the $\E^*$'s calculated is an upper bound. 

\begin{figure*}
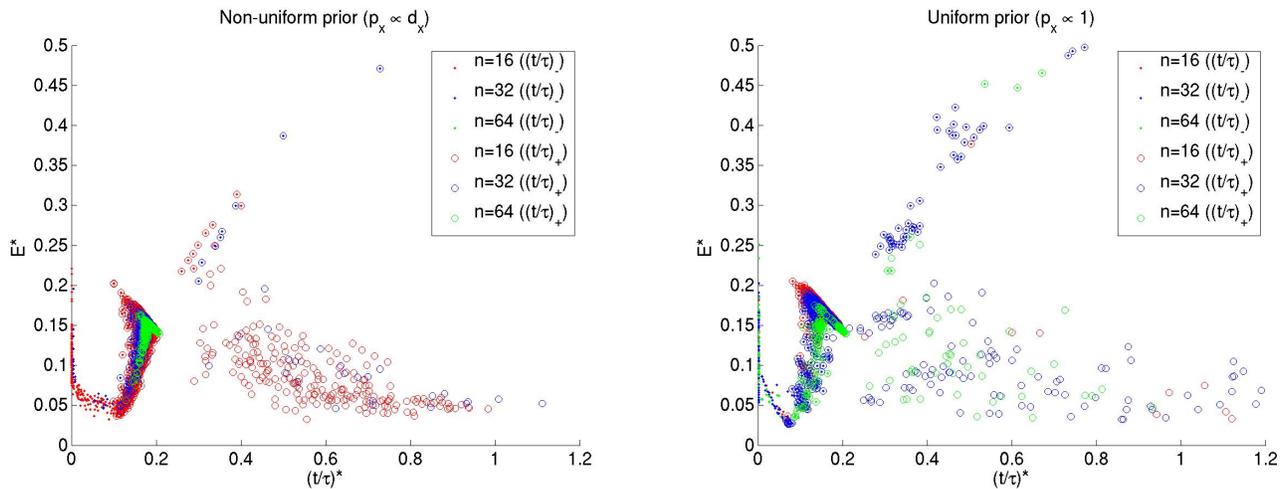

\centering
\begin{minipage}[t]{0.485\linewidth}
\spincgraph{plottwo0}
\end{minipage}
\quad
\begin{minipage}[t]{0.485\linewidth}
\spincgraph{plottwo1}
\end{minipage}
\caption{$\E^*$ vs $\ft^*$ for graphs of different sizes and different prior
distributions over the graph nodes. In the above plot, $\ft^*_-$ and $\ft^*_+$ are
represented by $\cdot$ and $\circ$, respectively.}
\label{plottwo}
\end{figure*}

Graphs were drawn randomly from a Stochastic Block Model (SBM) distribution \cite{sbm}, with block cardinality 2, 
to analyze the distribution of $\E^*$, $\ft^*_-$ and $\ft^*_+$. As is commonly done 
in community detection \cite{danon}, for a graph of $n$ nodes, 
the average degree per node is fixed at $n/4$ for graphs drawn from 
the SBM distribution: two nodes are connected with probability 
$p_{+}$ if they belong to the same block, but with probability $p_{-} < p_{+}$, 
if they belong to different blocks. The two probabilities are, 
thus, constrained by the relation
\beq
p_{+} \left( \frac{n}{2} - 1 \right) + p_{-} \left( \frac{n}{2} \right) = \frac{n}{4}
\eeq
leaving only one free parameter $p_{-}$ that tunes the `modularity' 
of graphs in the distribution. Starting with a graph drawn from a 
distribution specified by a $p_-$ value and specifying an initial cluster
assignment as given by the SBM distribution, we make local 
moves --- adding or deleting
an edge in the graph and/or reassigning a node's cluster label --- and  
search exhaustively over this move-set for local minima of $\E^*$. 
\fref{plottwo} 
compares the values of $\E^*$ and $\left \{ \ft^*_-,\ft^*_+ \right \}$ 
for graphs obtained in this systematic search,
starting with a graph drawn from a distribution with
$p_{-}=0.02$ and $n=\{16,32,64\}$. We note that the scatter plots for graphs 
of different sizes collapse on one another when $\E^*$ is plotted against 
normalized time, confirming the Fiedler value $1/\tau$ to be an 
appropriate characteristic diffusion time-scale as used in \cite{infomod}. 
A plot of $\E^*$ against actual diffusion time shows that the scatter plots 
of graphs of different sizes no longer collapse

\bfig
\cen
\incgraph{plotthree}
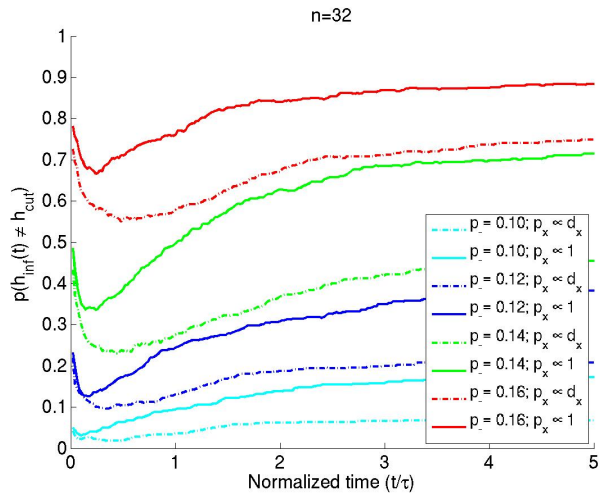
\caption{$p(\bh_{\rm{inf}}(t) \neq \bh_{\rm{cut}})$ vs normalized diffusion time,
averaged over 500 graphs drawn from a distribution parameterized by a given $p_{-}$ value,
is plotted for different graph distributions}
\label{plotthree}
\efig

Having shown analytically that for fast mixing graphs, the regularized mincut 
is approximately the rate of loss of relevance information,
it would be instructive to compare the actual 
partitions that optimize these goals. Graphs of size $n=32$ were drawn
from the SBM distribution with $p_{-}=\{0.1,0.12,0.14,0.16\}$. Starting 
with an equal-sized partition specified by the model itself, we performed 
iterative coordinate descent to search (independently) for the partition that 
minimized the regularized cut ($\bh_{\rm{cut}}$) and one that minimized 
the relevance information ($\bh_{\rm{inf}}(t)$); 
i.e. we reassigned each node's cluster label and searched for the reassignment 
that gave the new lowest value for the cost function being optimized. 
Plots comparing the partitions $\bh_{\rm{inf}}(t)$ and $\bh_{\rm{cut}}$, 
learnt by optimizing the two goals (averaged over 500 graphs drawn from 
each distribution), are shown in \fref{plotthree}.

\section{Concluding Remarks}

We have shown that the normalized cut and average cut, introduced by Shi and Malik as useful heuristics to be minimized 
when partitioning graphs, are well approximated by the rate of loss of predictive information for fast-mixing graphs. 
Deriving these cut-based cost functions from rate-distortion theory gives them a more principled setting, makes them 
interpretable, and facilitates generalization to appropriate cut-based cost functions in new problem settings. 
We have also shown (see \fref{plottwo}) that the
inverse Fiedler value is an appropriate normalization for diffusion time, justifying its use in \cite{infomod} to capture 
long-time behaviors on the network. 

Absent from this manuscript is a discussion of how not to overpartition a graph, 
i.e. a criterion for selecting K. It is hoped that by showing how these heuristics 
can be derived from a more general problem setting, lessons learnt by investigating 
stablilty, cross-validation or other approaches may benefit those using min-cut 
based approaches as well. Similarily, by showing how these heuristics approximate 
costs functions from a separate optimization problem, it is hoped that algorithms 
employed for rate distortion theory, e.g. Blahut Arimoto, maybe be brought to bear 
on min-cut minimization.

\renewcommand{\theequation}{A.\arabic{equation}}
\setcounter{equation}{0}  
\section*{APPENDIX}  
Using the definition of $\lap$, for any general vector $\bbf$ over the graph nodes, we have
\beq
\bbf^\bbT \lap \bbf & = & \bbf^\bbT \bD \bbf - \bbf^\bbT \bA \bbf\notag \\
& = & \sx d_x f_x^2 - \sxy f_x f_y A_{xy}\notag
\eeq
\beq
& = & \sx \left( \sy A_{xy} \right) f_x^2 - \sxy f_x f_y A_{xy}\notag \\
& = & \half \left( \sxy f_x^2 A_{xy} - 2 \sxy f_x f_y A_{xy}
+ \sxy f_y^2 A_{xy} \right)\notag \\
& = & \half \sxy A_{xy} \left( f_x - f_y \right)^2.
\eeq

Now, when $\bbf = \bh$, we have
\beq
\bh^\bbT \lap \bh & = & \half \shxym 4 A_{xy}\notag \\
& = & 4 \times c.
\eeq
The factor $\half$ disappears because summation over all nodes 
counts each adjacent pair of nodes twice. 

Using the definitions of $\A$ and $\N$, we have
\beq
\A & = & c \times \left( \frac{1}{\shxp 1} + \frac{1}{\shxm 1} \right) \notag \\
& = & c \times \left( \frac{1}{\sx \left( \frac{1+h_x}{2} \right) } 
+ \frac{1}{\sx \left( \frac{1-h_x}{2} \right) } \right) \notag \\
& = & 2 c \times \left( \frac{\sx (1-h_x+1+h_x)}{\sx (1+h_x) \sx (1-h_x)}  \right) \notag \\
& = & 2 c \times \left( \frac{2n}{(n+\sx h_x) (n-\sx h_x)} \right) \notag \\
& = & 2 c \times \left( \frac{2}{n(1+\bar{\bh})(1-\bar{\bh})} \right) \notag \\
& = & \frac{4}{n} \frac{c}{1-{\bar{\bh}}^2}.
\eeq
\beq
\N & = & c \times \left( \frac{1}{\shxp d_x} + \frac{1}{\shxm d_x} \right)\notag \\
& = & c \times \left( \frac{1}{\sx d_x \left( \frac{1+h_x}{2} \right) }
+ \frac{1}{\sx d_x \left( \frac{1-h_x}{2} \right) } \right)\notag \\
& = & 2 c \times \left( \frac{\sx d_x (1-h_x+1+h_x)}
{\sx \left( d_x (1+h_x) \right) \sx \left( d_x (1-h_x) \right)} \right)\notag \\
& = & 2 c \times \left( \frac{4m}{(2m+\sx h_x d_x) (2m-\sx h_x d_x)} \right)\notag \\
& = & 2 c \times \left( \frac{1}{m(1+\lhr)(1-\lhr)} \right)\notag \\
& = & \frac{2}{m} \frac{c}{1-\lhr^2}.
\eeq

\bibliographystyle{unsrt}
\bibliography{infospec}

\end{document}